\newif\ifconf
\newif\ifralfinal

\ralfinaltrue \conffalse

\ifralfinal
\documentclass[letterpaper, 10 pt, journal, twoside]{IEEEtran}
\fi
\ifconf
\documentclass[letterpaper, 10 pt]{IEEEtran}
\fi

\IEEEoverridecommandlockouts
\usepackage{cite}
\usepackage{amsmath,amssymb,amsfonts}
\usepackage{algorithmic}
\usepackage{graphicx}
\usepackage{textcomp}
\usepackage{xcolor}
\usepackage{balance}
\usepackage{comment}
\usepackage[ruled,vlined]{algorithm2e}
\usepackage{caption}
\usepackage{array}
\usepackage{subcaption}
\usepackage{placeins}
\usepackage{arydshln}
\usepackage{tensor}
\newcommand{\eg}{\emph{e.g.},}
\newcommand{\ie}{\emph{i.e.},}

\usepackage{hyperref}
\usepackage{float}
\usepackage{rotating}
\usepackage{nicematrix}  %
\usepackage{mathtools}
\usepackage{amsmath}
\usepackage{pifont}
\usepackage{cuted}
\usepackage{dblfloatfix}
\usepackage{multirow}

\usepackage{soul}

\begin{document}

\ifralfinal
\title{Pair-VPR: Place-Aware Pre-training and Contrastive Pair Classification for Visual Place Recognition with Vision Transformers
}
\fi
\ifconf
\title{\LARGE \bf Pair-VPR: Place-Aware Pre-training and Contrastive Pair Classification for Visual Place Recognition with Vision Transformers
}
\fi

\author{Stephen Hausler$^{1}$\textsuperscript{\textdagger}, Peyman Moghadam$^{1,2}$
\ifralfinal
\thanks{Manuscript received: Oct 3, 2024; Revised Jan 12, 2025; Accepted Feb 13, 2025. This paper was recommended for publication by Editor Sven Behnke upon evaluation of the Associate Editor and Reviewers' comments. This work was supported by the CSIRO's Data61 Embodied AI Cluster.}
\fi
\thanks{$^{1}$ CSIRO Robotics, Data61, CSIRO, Brisbane, QLD, Australia.
E-mail: {\tt\footnotesize \emph{peyman.moghadam}@csiro.au}}
\thanks{\textsuperscript{\textdagger} This work was done while the
author was at CSIRO.} 
\thanks{$^{2}$ School of Electrical Engineering and Robotics, Queensland University of Technology (QUT), Brisbane, Australia.}
}

\newcommand{\ourmethod}{\textit{Pair-VPR}}

\bstctlcite{IEEEexample:BSTcontrol}

\ifralfinal
\markboth{IEEE Robotics and Automation Letters. Preprint Version. Accepted February, 2025}{Hausler \MakeLowercase{\textit{et al.}}: Pair-VPR}
\fi

\maketitle

\ifralfinal
\begin{strip}
    \centering
    \vspace{-3.7cm} %
    \includegraphics[width=1.0\textwidth, trim=0cm 1.9cm 0cm 0.3cm, clip]{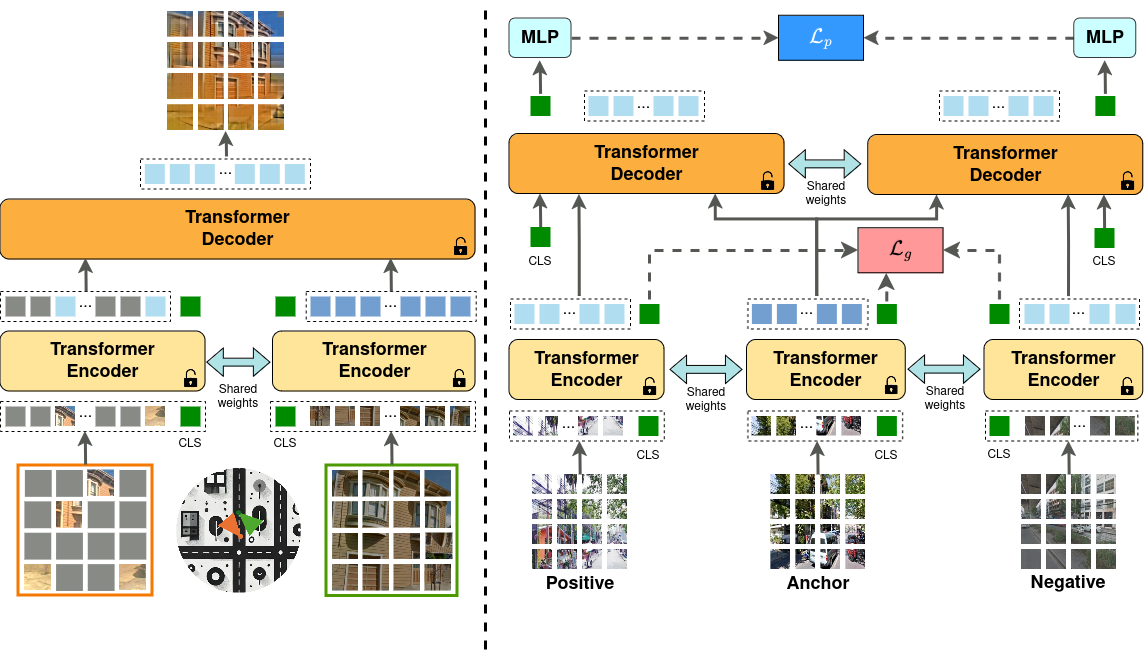}
    \captionof{figure}{Overview of the proposed \ourmethod{} method. \emph{Left:} In Stage 1 of training, we train a ViT encoder and decoder using Siamese mask image modelling with place-aware image sampling. \emph{Right:} In Stage 2, we re-use the pre-trained encoder and decoder and train specifically for the VPR task, jointly learning a global descriptor and a pair classifier.}
    \label{fig:hero}
    \vspace{-0.3cm}
\end{strip}
\fi

\begin{abstract}

In this work we propose a novel joint training method for Visual Place Recognition (VPR), which simultaneously learns a global descriptor and a pair classifier for re-ranking. The pair classifier can predict whether a given pair of images are from the same place or not. The network only comprises Vision Transformer components for both the encoder and the pair classifier, and both components are trained using their respective class tokens. In existing VPR methods, typically the network is initialized using pre-trained weights from a generic image dataset such as ImageNet. In this work we propose an alternative pre-training strategy, by using Siamese Masked Image Modeling as a pre-training task. We propose a Place-aware image sampling procedure from a collection of large VPR datasets for pre-training our model, to learn visual features tuned specifically for VPR. By re-using the Mask Image Modeling encoder and decoder weights in the second stage of training, \ourmethod{} can achieve state-of-the-art VPR performance across five benchmark datasets with a ViT-B encoder, along with further improvements in localization recall with larger encoders. The Pair-VPR website is: \href{https://csiro-robotics.github.io/Pair-VPR/}{https://csiro-robotics.github.io/Pair-VPR}

\end{abstract}

\ifralfinal
\fi

\section{Introduction}

\ifralfinal
\IEEEPARstart{T}{he}
\fi Place Recognition (PR) task involves associating input sensor data (\eg{} lidar\cite{vidanapathirana2022logg3d}, radar\cite{herraez2024spr}, or vision\cite{AR2018}), with a global map or database of places previously visited within an environment. In the field of learning-based Visual Place Recognition~(VPR)~\cite{AR2018,chen2017deep}, it is a standard approach to start with a pre-trained neural network, such as VGG16 or ResNet, leveraging their initial architecture and weights for the VPR task. Following this, it is common to incorporate a feature aggregation layer — such as VLAD~\cite{jegou2010aggregating}, GeM~\cite{radenovic2018fine}, Conv-AP~\cite{ali2022gsv}, SoP~\cite{vidanapathirana2022logg3d} among others — to pool local features into a compact global descriptor vector efficiently representing the original image. These vectors can then be compared with an embedding distance metric such as the Euclidean distance, with the smallest distance noting the pair of most similar images. More recent VPR techniques often include multiple stages of retrieval, where subsequent stages of retrieval are used to re-rank an initial set of candidate place recognition matches~\cite{lu2024towards, hausler2021patch, wang2022transvpr, vidanapathirana2023spectral}.

In this work, we develop \ourmethod{}, a transformer-based VPR method trained on diverse VPR datasets, which achieves \emph{state-of-the-art} visual place recognition performance across a range of challenging VPR benchmarks. We achieve this by proposing a \emph{two-stage} training pipeline to train a \emph{pair-classifier}, which can decide whether a given pair of images is from the same place or not. In the first stage, we pre-train a transformer encoder and decoder using Siamese mask image modeling~\cite{gupta2023siamese, weinzaepfel2022croco}, with place-aware image sampling. We sample pairs of images from different places in the world, ensuring that these pairs contain both \emph{spatial} and \emph{temporal} differences. Our places are curated from three existing large-scale open source datasets (SF-XL~\cite{berton2022rethinking}, GSV-Cities~\cite{ali2022gsv} and Google Landmarks v2~\cite{weyand2020google}) with a total of $3.43$ million panoramic images and $2.08$ million egocentric images, across diverse locations (planet-wide) and diverse times.

In the second stage, we re-use both the encoder and decoder, by jointly learning to produce a global descriptor from the encoder and using the decoder as a pair classification network. The pair classifier produces a similarity score denoting whether a given pair of images were captured from the same location, or not. A diagram showing our network architecture and training recipe is shown in Figure 1. The \ourmethod{} network uses a vision transformer setup, with ViT blocks in both the encoder and decoder. We leverage the class token output to supervise the network, with a low-dimensional global descriptor produced from an encoder class token and a pair similarity score from decoder class tokens. The network is then trained using a Multi-Similarity loss for the global descriptor and a Binary Cross-Entropy loss for the pair classifier, with online triplet mining. We benchmark the trained network on existing VPR benchmark datasets and observe state-of-the-art Recall@1 on all tested datasets with a ViT-B encoder, such as an improvement in the Recall@1 on the Tokyo24/7 dataset from $95\%$ to $98\%$. Moreover, we show that our proposed method can be extended to much larger vision transformer encoders (ViT-L, ViT-G) and achieve a new benchmark result of Recall@1 on Tokyo24/7 of $100\%$ with our best performing configuration.

\section{Related Work}

\subsection{Visual Place Recognition}

Compared to the general task of Place Recognition, VPR focuses only on the image modality and this allows for VPR to be considered as an image retrieval problem. This formulation in conjunction with the advent of deep learning led to a number of high performing learning-based VPR solutions. Initially, pre-trained neural networks were applied to VPR, then NetVLAD was developed which utilized triplet loss to train a neural network specifically for the place recognition task~\cite{AR2018}. Numerous subsequent works expanded upon NetVLAD, including works that relax the triplet loss to use soft positives and negatives~\cite{thoma2020soft}, quadruplets rather than triplets~\cite{uy2018pointnetvlad}, a generalized contrastive loss~\cite{leyva2023data}, and aggregates features over multiple spatial scales~\cite{yu2019spatial}. Other learning-based methods include those which consider place recognition as a classification problem~\cite{berton2022rethinking,chen2017deep} or use a multi-similarity loss~\cite{ali2023mixvpr}. These earlier works all follow the same approach, of constructing a \emph{global descriptor}, \ie{} a single vector representing an image for nearest neighbor retrieval. However, the top candidate found using global descriptor matching is not always correct. To avoid such limitations, multi-stage VPR algorithms propose to re-rank a collection of top candidates.

These \emph{re-ranking} methods involve re-ordering an initial set of retrieved places such that the correct place match is ranked first in the retrieved list. 
Learning-based \emph{re-ranking} methods began with less computationally efficient algorithms that used geometric verification densely (without keypoint selection)~\cite{hausler2021patch}, then subsequently evolved to include keypoint/keypatch selection algorithms~\cite{li2023hot, wang2022transvpr}. More recently, R2Former~\cite{zhu2023r2former} went one step further by removing explicit geometric verification altogether. Instead, they treat re-ranking as a classification problem and use a small two-block transformer for selecting the best candidate image. Similarly, GeoAdapt~\cite{geoadapt2024} also learns a classifier except instead for the task of classifying the similarity of two point clouds for LiDAR Place Recognition. In our proposed approach, we show that a re-ranking classification transformer can be designed using standard self and cross-attention blocks and scaled up to over $80$ million parameters, by leveraging a VPR-specific pre-training recipe.

\subsection{Vision Transformers and Self-Supervised Learning}

Compared to Convolutional Networks (CNNs), Vision Transformers~\cite{dosovitskiy2020image} have no inductive bias therefore they can learn global dependencies between image features without being limited by a specific kernel size; however, as a result, larger image datasets are needed to train transformer models. This led to the concept of Self-Supervised Learning (SSL) techniques for pre-training these large transformer models on large quantities of unlabeled data. These techniques include deep metric learning methods such as SimCLR~\cite{chen2020simple}, BYOL~\cite{grill2020bootstrap}, and DINO~\cite{caron2021emerging}. The recent work DINOv2~\cite{oquab2023dinov2} provided further improvements to the pre-training recipe and in this work, we leverage the robust learnt features they provide. In contrast to metric learning methods, an alternate self-supervised approach is called Masked Image Modeling (MIM) and will be explained in the following subsection.

\subsection{Masked Image Modeling (MIM) and Siamese MIM}

Masked Image Modeling is a SSL technique where a neural network is tasked with reconstructing a partially masked image, where the image is divided into a set of non-overlapping patches. Initially, this reconstruction was performed on a set of transformer tokens produced from a raw image~\cite{bao2021beit}. Subsequently, MAE~\cite{he2022masked} demonstrated that instead the reconstruction could be performed directly on the raw pixel values, based on the mean squared pixel error between the original and reconstructed image. Further works then expanded upon this concept via improvements such as distillation (iBOT~\cite{zhou2021ibot}), continuous masking (ROPIM~\cite{haghighat2024ropim}), and cross-view completion (CroCo~\cite{weinzaepfel2022croco}). In CroCo, pairs of images are provided to the network where one image is masked and a second image, showing the same scene, is provided to the network without any masking. They showed that this pre-training recipe is ideal for 3D vision tasks. In a concurrent work, a similar architecture was proposed for learning from video data, called Siamese Masked Autoencoders~\cite{gupta2023siamese}. In our approach, we also provide pairs of images to a Mask Image Modeling network, except we propose a \emph{Place-aware} training methodology for VPR.

\section{Methodology}

Our proposed Visual Place Recognition (\ourmethod{}) solution is a two-stage place recognition method, which first uses a global descriptor to generate a list of potential place matches, then uses a second stage to refine these matches to improve the likelihood that the highest ranked match is correct. For the first time, we propose a two-stage training methodology using a mask image modeling pre-training designed specifically for VPR. Then we propose a second stage approach where we learn to generate both a global descriptor and a pair classifier for VPR.

During the first stage, we provide pairs of images to the network and then heavily mask one of these images. The training objective is to reconstruct the masked image, using a network that has both encoder and decoder components. In the second stage we jointly optimize both the encoder and decoder for VPR, generating a global descriptor from the class token of the encoder, and the decoder is trained to predict whether a given pair of images is a positive or negative pair.

After training, the network can then be used as a two-stage VPR method. The encoder is used to produce low dimensional global descriptors which are then used for nearest neighbor searching through a VPR database. Then the decoder is used to decide which potential database candidate is the best match for the current query by passing (query, database) pairs to the decoder. An overview of our pre-training and fine-tuning procedures are provided in Figure~\ref{fig:hero}.

\subsection{Stage One: Place-Aware Masked Image Modeling}

In the first stage of training, we use the Siamese Masked Autoencoder~\cite{gupta2023siamese} design to train the \ourmethod{} network. In this approach, pairs of images are input into the network, and one of the images is heavily masked. The network is then able to leverage the visual information in the second, unmasked image to aid in reconstructing the masked first image. In previous works, generating pairs is achieved either by sampling random frames in a video sequence~\cite{gupta2023siamese}, or by sampling two different viewpoints of a scene\cite{weinzaepfel2022croco}. In our approach, we propose a location sampling strategy, such that pairs are sampled from specific locations and, in VPR terminology, are always collected from the set of \emph{positives} from a given place.

\vspace{0.7em}
\noindent \textbf{Network:}
Our network comprises a shared encoder that converts a first image $I_a$ (masked) and a second image $I_b$ (unmasked) into latent representations. We use a Vision Transformer~\cite{dosovitskiy2020image} encoder and convert each image into a collection of non-overlapping patches, which are then converted to a set of tokens. For the masked image, we replace the majority of image tokens with mask tokens, which are learnt parameters but have no information received from the original image.

After the encoder, we pass the set of encoded patch embeddings (excluding any class tokens) from both images through a decoder to reconstruct the masked image. Our decoder consists of another vision transformer except with alternating self-attention and cross-attention layers, with self-attention between tokens from $I_a$ and cross-attention between tokens from $I_a$ and $I_b$. For efficiency, we use Flash Attention~\cite{dao2022flashattention}. After a number of decoder blocks, we pass the features through a final FC layer to produce reconstructed pixel values per token (\ie{} per image patch).

\vspace{0.7em}
\noindent \textbf{Loss:}
The network is trained using a reconstruction loss between the predicted and ground truth (unmasked) images, by minimizing the Mean Squared Error between predicted and true pixel values. The loss function is expressed as below:
\begin{equation}
    \mathcal{L} \left(I_a,I_b \right) = \frac{1}{\vert p_a / \tilde{p}_a \vert} \sum_{p_a^i \in p_a/\tilde{p}_a} \vert\vert \hat{p}_a^i - p_a^i \vert\vert ^2 , 
\end{equation}
where $I_a$ and $I_b$ denote the two input images, $p_a$ denotes the ground truth value for a masked pixel at index $i$ from image $I_a$, $\hat{p}_a$ denotes the predicted pixel value, and $p_a/\tilde{p}_a$ is the subset of masked pixels from the masked image. The loss function is only calculated for any pixels that have been masked, where the mask ratio is a hyperparameter of the network. Before calculating the loss we normalize each patch by the mean and standard deviation of that patch as per prior work~\cite{he2022masked}.

\vspace{0.7em}
\noindent \textbf{Training:}
Our network is pre-trained in a \emph{Place-aware} fashion, where each iteration is a defined physical location in the world. Then pairs of images are sampled from this place, with dataset-specific sampling to ensure that sampled pairs have some viewpoint consistency; it would be almost impossible for the network to reconstruct an image using a second image facing the opposite direction. Our training data can theoretically be drawn from any collection of places in the world, however, in this work we limit ourselves to a set of pre-existing open-source datasets. Further details concerning dataset specific implementations are provided in Section Four.

\subsection{Stage Two: Contrastive Pair Classification for VPR}

In the second stage, the encoder is trained to generate global descriptors for retrieval, and the decoder is trained to predict whether a given pair of images is similar or not. We used the pre-trained weights from stage one of \ourmethod{}, loading weights from \emph{both} the encoder and the mask image modeling decoder. We then \emph{jointly} train the encoder and decoder for both global retrieval and pair classification simultaneously. Finally, we use class tokens from both the encoder and decoder as inputs to our VPR loss.

\vspace{0.7em}
\noindent \textbf{Network:}
The stage two network architecture is an extension to the stage one network for the VPR task. We add a linear layer to project from the class token in the encoder to a smaller dimension, then use an $L_2$ norm to generate a global descriptor. We then convert the Mask Image Modeling decoder into a pair classifier by adding a new class token and adding a two-layer MLP to the output of this class token to produce a scalar value denoting the similarity between a pair of images. The network architecture of our \ourmethod{} is shown in Figure~\ref{fig:hero}.

\vspace{0.7em}
\noindent \textbf{Loss:}
We use a contrastive loss to train the encoder for the global descriptor and a Binary Cross Entropy (BCE) loss to train the decoder for pair classification. We use online mining and use a Multi-Similarity Miner~\cite{Musgrave2020PyTorchML} to select positives, negatives, and anchors before using a Multi-Similarity Loss~\cite{wang2019multi} for the global descriptor, with the loss function shown below: 

\begin{multline}
    \mathcal{L}_{g} = \frac{1}{N} \sum_{i=1}^{N}  \big(  \frac{1}{\alpha} \log \left[ 1 + \sum_{k \in \mathcal{P}_{i}} e^{- \alpha \left(S_{ik} - m \right)} \right] \\ +    \frac{1}{\beta}  \log \left[ 1+\sum_{k \in \mathcal{N}_{i}} e^{\beta \left(S_{ik} - m \right)} \right] \big), %
\end{multline}
where $N$ is the batch size and $S_{ik}$ is the similarity value between the index $i$ in the batch and a positive/negative index $k$. We keep the hyperparameters $\alpha, \beta, m$ the same as used in~\cite{ali2022gsv}.

During the forward pass, we store both the global descriptor and the dense features from the encoder and use the dense features as input into the decoder. As the decoder requires an image pair, the challenge is to provide examples to the network of both positives (two images of the same place) and negatives (two images from different places). To prevent the network from converging to the trivial solution, we use an Online Hardest Batch Negative Miner~\cite{Musgrave2020PyTorchML} to use the hardest positives, anchors, and negatives in a given batch. 

We then pass into the decoder sets of (anchor, positive) pairs and (anchor, negative) pairs to produce a list of scalar values per pair, where a large scalar value denotes a high similarity between a given pair. After concatenating both sets, we use a BCE loss (with a sigmoid function) to optimize the network with positive pairs having a target value of $1$ and negative pairs having a target value of $0$:
\begin{equation}
    \mathcal{L}_p = -\frac{1}{N} \sum_{n=1}^{N}\big(y_n \log s_n+(1-y_n)\log(1-s_n)\big),
\end{equation}
where $y_n$ is the target for the current pair ($I_a, I_b$) and $s_n$ is the similarity output for the pair for a batch size of $N$ pairs. We then train the network jointly, with the full loss shown below:
\begin{equation}
    \mathcal{L} = \mathcal{L}_g + w \mathcal{L}_p,
\end{equation}
where $L_g$ is the global loss trained using a Multi-Similarity loss and miner~\cite{Musgrave2020PyTorchML, ali2023mixvpr} and $L_p$ is the BCE pair loss. $w$ is a hyperparameter to balance the two loss terms.

\subsection{Using \ourmethod{} during Inference}

Once \ourmethod{} is trained, VPR is performed using a two-stage process. First, all images are passed through the encoder and global descriptors are generated. These global descriptors are then used to find the top $N$ database candidates for each query. We also save dense features for second-stage refinement - for an image input of size $322$ by $322$ pixels, these have a size of $529$ by $768$ with a ViT-B encoder.

\begin{figure}[t]
    \centering
    \includegraphics[width=0.85\columnwidth, trim=0.8cm 16.0cm 8.4cm 1.0cm, clip]{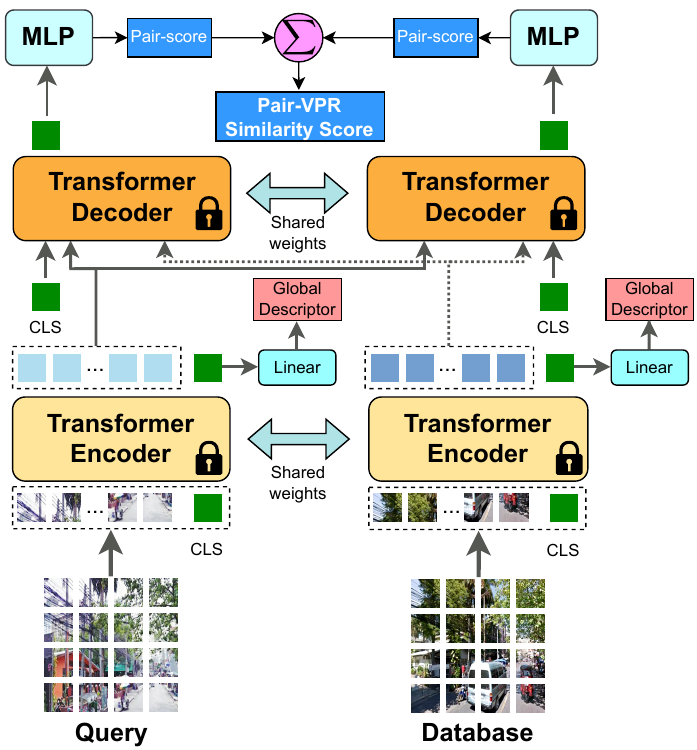}
    \caption{During inference, we pass in pairs of (query, database) images and the network produces a score estimating whether or not the pair of images are from the same location or not, along with a global descriptor per image.}
    \label{fig:eval}
    \vspace{-0.3cm}
\end{figure}

In the second stage, for each query, we copy the dense features by $N$ in order to create batches of (query, database$_j$) pairs, where $j$ is the database index and there are $N$ pairs in total. Because our decoder is not symmetric (self-attention is only performed on the first image in a pair), we pass into the decoder batches of both (query, database$_j$) and (database$_j$, query) pairs and then sum their respective scores. The highest-scoring pair is then considered the best match for the current query. Figure~\ref{fig:eval} shows our network during evaluation.

\section{Implementation Details}

\subsection{Stage One Training}

We begin by pre-initializing the ViT encoder with weights from DINOv2~\cite{oquab2023dinov2}. We found that leveraging the diverse pre-training policy used in DINOv2 improved performance over random initialization (please refer to ablation studies in Table~\ref{tab:abl}). We then freeze the first half of the encoder blocks (\eg{} six blocks with ViT-B) and train the second half using Place-Aware Masked Image Modeling. We train in a Place-aware fashion and construct a dataloader such that a single item in a batch is a single \emph{place}, where an item comprises a pair of images. We selected three existing large open source datasets for stage one training: SF-XL~\cite{berton2022rethinking}, GSV-Cities~\cite{ali2022gsv}, and Google Landmarks v2~\cite{weyand2020google}. 

\vspace{0.7em}
\noindent \textbf{SF-XL:}
We follow the procedure described in EigenPlaces\cite{berton2023eigenplaces} and divide the dataset into $M \times M$ meter-sized cells based on the UTM coordinates of all panoramic images in the dataset, generating a total of $C$ cells. Considering the $i$-th cell $C_i$, we follow the approach in EigenPlaces of computing the Singular Value Decomposition (SVD) from the UTM coordinates of all images in this cell, along with the mean UTM coordinate $\mu_i$ for this cell. We then select pairs of panoramic images randomly from each cell. Given cell $C_i$, we begin by calculating a random focal point (\eg{} a target UTM coordinate) that we want our pairs to observe, anchored to the first principle component of the cell denoted as $V_0^i$. Our formula for calculating a focal point is given below:
\begin{equation}
    f_i = \mu_{i} + D_i \times \textbf{R}\left( V_0^i, O_i\right),
\end{equation}
where $O_i$ is a random observation angle, and $D_i$ is a random focal length. At each iteration, we randomly sample observation angles $O_i$ between $0$ and $360$ degrees and apply this random rotation to the eigenvector $V_0^i$. We then randomly sample different focal lengths $D_i$, sampling between 10 and 20 meters away from the mean coordinate $\mu_i$. This approach differs from the method in EigenPlaces, which only utilized the $0$ and $90$ degree observation angles and a fixed focal length.

Then given a focal point $f_i$, we produce crops from two randomly sampled panoramic images within cell $C_i$, such that the cropped views are focused on the target focal point - this ensures that our pair of images contains overlapping visual information, without requiring any manual curation of pairs. We can then calculate a viewing angle between a given panoramic image $j$ and a focal point using their respective UTM coordinates:
\begin{equation}
    \alpha_j = arctan \left( \frac{e_f - e_j}{n_f - n_j} \right).
\end{equation}

Given this viewing angle, we select a $512$ by $512$ pixel crop from a $3328$ by $512$ pixel panoramic image. However, pairs can be selected that are either too easy (small viewpoint difference), or too hard (too much viewpoint difference). Therefore, we calculate the difference in angles between our two sampled crops:

\begin{equation}
    \theta_i = \angle \alpha_{1} - \angle \alpha_{2}.
\end{equation}

Then we check if $\theta_i$ is between a low and high threshold $3$\textdegree and $50$\textdegree. If not, we resample until we return a pair of crops within the required range.

\vspace{0.7em}
\noindent \textbf{GSV-Cities:}
We randomly sample a pair of images from each place in the GSV-Cities dataset, excluding places with less than $4$ images available.

\vspace{0.7em}
\noindent \textbf{GLDv2:}
To improve the data diversity, we also added the Google Landmarks dataset~\cite{weyand2020google}. We only use the cleaned subset of the dataset to avoid damaging the network's ability to learn by providing ambiguous image pairs. Additionally, we also exclude any landmarks with less than $4$ images. We consider a landmark as a proxy for a \emph{place} and treat the dataset in the same format as SF-XL and GSV-Cities. In total, we are left with $72,322$ landmarks for training after cleaning and filtering.

\vspace{0.7em}
\noindent \textbf{Training Summary:}
We merge the three aforementioned datasets to produce a total of $266809$ \emph{places} per epoch, which are sampled from a set of $3.43$ million panoramic images and $2.08$ million egocentric images. We train using a mask ratio of $90$ percent with a ViT-B sized decoder for $500$ epochs with a learning rate of {2e-4} for a batch size of $512$ places with AdamW and a cosine LR scheduler. We train using square images of size $224$ pixels. By default, we use a ViT-B encoder with $4$ register tokens, however, we note that our approach can also work with larger encoder sizes. Hyperparameters for larger encoders are kept the same except we use 1000 epochs. 

\subsection{Stage Two Training}

We continue training our model using the GSV-Cities dataset~\cite{ali2022gsv} following the training recipe used in prior works such as MixVPR~\cite{ali2023mixvpr} and SALAD~\cite{izquierdo2023optimal}. We train for $10$ epochs using a linear LR scheduler with an initial LR of {8e-5}, with a weight decay of {5e-2} and a batch size of $100$ places. We also use square images of size $322$, and freeze all encoder layers except the last six. Our global descriptor has only $512$ dimensions and we set $w$ (the loss balance term) to $2$. Hyperparameters were set using a grid search on the MSLS-Val set. We use MSLS-Val as our validation dataset and take the checkpoint with the highest second-stage Recall@1 as the final trained model.

\subsection{Evaluation}

We evaluate our method on five commonly used benchmark VPR datasets, with a diverse set of environments. The datasets are: MSLS validation set~\cite{warburg2020mapillary}, MSLS challenge set~\cite{warburg2020mapillary}, Pittsburgh30k~\cite{torii2013visual}, Tokyo247~\cite{torii201524} and Nordland~\cite{skrede2013nordlandsbanen} (note that we use the split of Nordland from VPRBench~\cite{zaffar2021vpr}, with $2760$ query images). During the evaluation, we resize all images to $322\times322$ resolution. By default, Pair-VPR uses a ViT-B encoder and uses the top 100 candidates after global descriptor matching - we refer to this as the Speed configuration of \ourmethod{} (Pair-VPR-s). We also provide a Performance version of \ourmethod{}, which has a ViT-G encoder and uses the top $500$ candidates during pair-classification (Pair-VPR-p). We evaluate using the standard Recall@N metric~\cite{AR2018,berton2022rethinking} with $N \in 1,5,10$.

To compare \ourmethod{} with existing techniques, we benchmark against five existing single-stage learnt VPR methods (CosPlace~\cite{berton2022rethinking}, MixVPR~\cite{ali2023mixvpr}, EigenPlaces~\cite{berton2023eigenplaces}, BoQ~\cite{ali2024boq} and SALAD~\cite{izquierdo2023optimal}) and three existing two-stage VPR techniques (Patch-NetVLAD~\cite{hausler2021patch}, R2Former~\cite{zhu2023r2former} and SelaVPR~\cite{lu2024towards}). We also include a baseline single-stage ViT VPR system we refer to as PaVPR (Place-aware VPR), which uses the same Stage One training as Pair-VPR except discards the pair contrastive training component during Stage Two and has a global descriptor with 512 dimensions.

\section{Results and Discussion}

\newcount\columncount
\columncount = 17

\begin{table*}[!t]
  \scriptsize
  \setlength\tabcolsep{0.1cm}
  \renewcommand{\arraystretch}{1.2}
  \centering
  \caption{Quantitative results for VPR methods on benchmark datasets. Rows above and below the line denote single-stage and two-stage VPR methods respectively. Underlined numbers denote the second best in a column.} %
  \resizebox{0.99\linewidth}{!}{
    \begin{tabular}{c|c|ccc||ccc||ccc||ccc||ccc}
\multirow{2}{*}{Method} & \multirow{2}{*}{Enc. Size} & \multicolumn{3}{c||}{\textbf{MSLS-Val}}& \multicolumn{3}{c||}{\textbf{MSLS-Challenge}}& \multicolumn{3}{c||}{\textbf{Pitts30k}}& \multicolumn{3}{c||}{\textbf{Tokyo24/7}}& \multicolumn{3}{c}{\textbf{Nordland}} \\
\cline{3-\columncount}
& & R@1 & R@5 & R@10 & R@1 & R@5 & R@10 & R@1 & R@5 & R@10 & R@1 & R@5 & R@10 & R@1 & R@5 & R@10 \\
\cline{1-\columncount}
\noalign{\vskip\doublerulesep\vskip-\arrayrulewidth}
\cline{1-\columncount}
CosPlace~\cite{berton2022rethinking} & ResNet-50 & 87.4 & 94.1 & 94.9 & 67.0 & 77.9 & 80.7 & 90.9 & 95.7 & 96.7 & 87.3 & 94.0 & 95.6 & 55.3 & 71.5 & 77.5 \\
MixVPR~\cite{ali2023mixvpr} & ResNet-50 & 88.0 & 92.7 & 94.6 & 64.0 & 75.9 & 80.6 & 91.5 & 95.5 & 96.3 & 85.1 & 91.7 & 94.3 & 58.4 & 74.6 & 80.0 \\
EigenPlaces~\cite{berton2023eigenplaces} & ResNet-50 & 89.1 & 93.8 & 95.0 & 67.4 & 77.1 & 81.7 & 92.5 & 96.8 & 97.6 & 92.4 & 96.2 & 97.1 & 54.4 & 68.8 & 74.1 \\
BoQ~\cite{ali2024boq} & ResNet-50 & 91.2 & 95.3 & 96.1 & 72.8 & 83.1 & 86.3 & 92.4 & 96.5 & 97.4 & 90.5 & 95.2 & 96.5 & 70.7 & 84.0 & 87.5\\
SALAD~\cite{izquierdo2023optimal} & ViT-B & 92.2 & 96.4 & 97.0 & 75.0 & \underline{88.8} & \textbf{91.3} & 92.4 & 96.3 & 97.4 & 95.2 & 97.1 & 98.1 & 76.0 & 89.2 & \underline{92.0}\\
PaVPR baseline & ViT-B & 89.5 & 94.5 & 95.8 & 67.8 & 80.5 & 83.8 & 90.4 & 95.4 & 96.8 & 86.4 & 94.6 & 95.6 & 57.7 & 75.5 & 80.8\\ 
\cline{1-\columncount} %
Patch-NetVLAD-s~\cite{hausler2021patch} & VGG-16 & 77.3 & 84.2 & 86.5 & 35.8 & 50.1 & 55.3 & 87.1 & 93.9 & 95.4 & 67.6 & 74.6 & 77.8 & 24.8 & 29.2 & 30.7 \\
R2Former~\cite{zhu2023r2former} & ViT-S & 89.7 & 95.0 & 96.2 & 73.0 & 85.9 & 88.8 & 91.1 & 95.2 & 96.3 & 88.6 & 91.4 & 91.7 & - & - & - \\
SelaVPR~\cite{lu2024towards} & ViT-L & 90.8 & 96.4 & 97.2 & 73.5 & 87.5 & \underline{90.6} & 92.8 & 96.8 & 97.7 & 94.0 & 96.8 & 97.5 & 66.2 & 79.8 & 84.1 \\
\textbf{Pair-VPR-s (Ours)} & ViT-B & \underline{93.7} & \underline{97.2} & \underline{97.3} & \underline{79.0} & 86.9 & 88.3 & \underline{94.7} & \underline{97.2} & \underline{97.8} & \underline{98.1} & \underline{98.4} & \underline{98.7} & \underline{84.2} & \underline{90.9} & 91.6 \\
\textbf{Pair-VPR-p (Ours)} & ViT-G & \textbf{95.4} & \textbf{97.3} & \textbf{97.7} & \textbf{81.7} & \textbf{90.2} & \textbf{91.3} & \textbf{95.4} & \textbf{97.5} & \textbf{98.0} & \textbf{100} & \textbf{100} & \textbf{100} & \textbf{91.0} & \textbf{95.2} & \textbf{95.7} \\
\end{tabular}%
    }
  \label{tab:results}%
  \vspace*{-0.2cm}
\end{table*}%

\newcount\columncount
\columncount = 15

\begin{table*}[!t]
  \scriptsize
  \setlength\tabcolsep{0.1cm}
  \renewcommand{\arraystretch}{1.2}
  \centering
  \caption{Stage One Training Ablation Study for VPR. Bold denotes the standard configurations of \ourmethod{}.}%
  \resizebox{0.99\linewidth}{!}{
    \begin{tabular}{ccc|ccc||ccc||ccc||ccc}
\multirow{2}{*}{Ablation} & \multirow{2}{*}{Enc. size} & \multirow{2}{*}{Initialization} & \multicolumn{3}{c||}{\textbf{MSLS-Val}} & \multicolumn{3}{c||}{\textbf{Pitts30k}} & \multicolumn{3}{c||}{\textbf{Tokyo24/7}} & \multicolumn{3}{c}{\textbf{Nordland}}\\
\cline{4-\columncount}
& & & R@1 & R@5 & R@10 & R@1 & R@5 & R@10 & R@1 & R@5 & R@10 & R@1 & R@5 & R@10 \\
\cline{1-\columncount}
\noalign{\vskip\doublerulesep\vskip-\arrayrulewidth}
\cline{1-\columncount}

No stage one training & ViT-B & DINOv2 & 78.2 & 95.4 & 96.1 & 89.3 & 95.3 & 96.5 & 92.1 & 97.1 & 97.8 & 54.5 & 74.6 & 78.6 \\

Training without pairs & ViT-B & DINOv2 & 91.2 & 95.0 & 95.7 & 91.8 & 95.8 & 96.8 & 93.0 & 94.9 & 95.6 & 67.3 & 76.4 & 78.5 \\

Unfreeze all encoder blocks & ViT-B & DINOv2 & 91.9 & 96.0 & 96.8 & 93.1 & 96.5 & 97.2 & 94.3 & 95.9 & 96.2 & 78.7 & 86.6 & 88.3 \\

Train from scratch & ViT-B & Random & 88.4 & 91.5 & 91.8 & 92.8 & 95.8 & 96.4 & 83.2 & 85.4 & 86.4 & 42.7 & 45.9 & 46.3 \\

\cline{1-\columncount}

\textbf{Pair-VPR-s} (Top100-B) & ViT-B & DINOv2  & 93.7 & \underline{97.2} & 97.3 & \underline{94.7} & \underline{97.2} & 97.8 & 98.1 & \underline{98.4} & \underline{98.7} & 84.2 & 90.9 & \underline{91.6} \\

Pair-VPR (Top100-L) & ViT-L & DINOv2 & 94.5 & 97.0 & 97.4 & 94.0 & 97.0 & 97.8 & 96.8 & 96.8 & 96.8 & \underline{86.6} & \underline{91.2} & 91.5 \\

Pair-VPR (Top100-G) & ViT-G & DINOv2 & \underline{95.1} & \underline{97.2} & \underline{97.6} & \textbf{95.4} & \textbf{97.5} & \underline{97.9} & \underline{98.4} & \underline{98.4} & 98.4 & 85.8 & 89.1 & 89.4 \\

\textbf{Pair-VPR-p} (Top500-G) & ViT-G & DINOv2 & \textbf{95.4} & \textbf{97.3} & \textbf{97.7} & \textbf{95.4} & \textbf{97.5} & \textbf{98.0} & \textbf{100} & \textbf{100} & \textbf{100} & \textbf{91.0} & \textbf{95.2} & \textbf{95.7} \\

\end{tabular}%
    }
  \label{tab:abl}%
  \vspace*{-0.4cm}
\end{table*}

\subsection{Comparison to Recent VPR Methods}

We begin by comparing the performance of \ourmethod{} against other State-Of-The-Art (SOTA) VPR methods on a range of benchmark datasets (Table~\ref{tab:results}). Comparing our method, Pair-VPR-s, to the other methods we benchmark, we observe that \ourmethod{} achieves the highest Recall@1 on all five datasets. We observe that our transformer-based re-ranking network is particularly adept at improving the Recall@1 over other works that also use a vision transformer backbone (SALAD~\cite{izquierdo2023optimal}, SelaVPR~\cite{lu2024towards}, and our PaVPR baseline). Comparing against R2Former~\cite{zhu2023r2former}, a prior SOTA method that also used a re-ranking/pair-classifier transformer, we observe a significant jump in performance, especially on the Tokyo dataset.

When we scale up our method to use larger encoder sizes and more top candidates, we observe that Pair-VPR-p provides a large performance increase over Pair-VPR-s and achieves the highest recall across all Recall@N values compared to prior works. We especially highlight the results on Tokyo24/7, where we have achieved a Recall@1 of $100$\%.

\begin{figure}[h]
    \centering
    \includegraphics[width=0.85\linewidth, trim=0cm 0.2cm 0cm 0.2cm, clip]{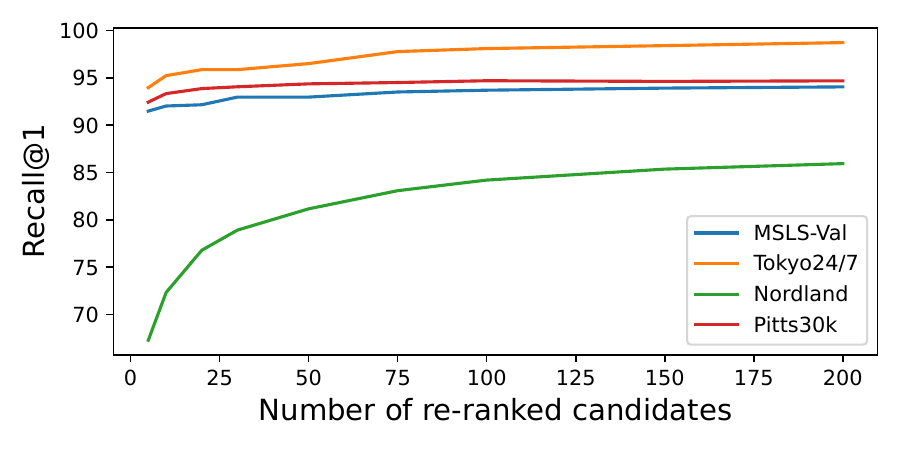}
    \caption{Recall@1 as the number of re-ranked candidates is increased for Pair-VPR-s.}
    \label{fig:topn}
    \vspace{-0.4cm}
\end{figure}

We also include results on Pair-VPR-s when the number of top candidates that are re-ranked by Pair-VPR are varied (Figure~\ref{fig:topn}). The graph shows that a lower number of top candidates can still achieve good VPR performance, especially on the MSLS-Val and Pitts30k datasets. This is because the Pair-VPR global descriptor has good performance on these datasets. Tokyo24/7 exhibits a similar trend, except that a larger number of top candidates allows the Pair-VPR re-ranking to achieve very high recalls. We observe that increasing the number of top candidates has significant benefits on the Nordland dataset. The Nordland split from VPRBench is very challenging, with a lot of perceptual aliasing and a small localization error tolerance; therefore, a larger number of top candidates is necessary on this dataset.

\subsection{Pre-training Ablation Study}

To understand which aspects of our training strategy are essential for the performance of \ourmethod{}, we conducted an ablation study over different variations of stage one training (Table~\ref{tab:abl}). In the first row, we show the results when we do not perform stage one training at all, with the decoder in stage two initialized using random weights. We observed that pre-training the decoder network via stage one training is essential for achieving effective VPR performance - since the pair-classification task is non-trivial, the place-aware Siamese mask image modeling task provides an initial set of weights that are already tuned for comparing features between two images.

In the second row of Table II, we investigated the importance of place-aware sampling by removing the place sampling during stage one training, and instead used strong augmentation to generate the second unmasked image. We observed a consistent drop in recall, even though the total collection of training images remains identical.

In row three, we experimented with unfreezing all blocks in our DINOv2-initialized ViT encoder. We found that allowing the entire network to be trained reduced the VPR performance, and we hypothesize that this is because the DINOv2 network was originally trained on a larger and more diverse dataset than ours, and maintaining the low level (\eg{} color) learnt features from a more diverse dataset improves performance.

In the fourth row we instead initialized the ViT encoder with random weights and performed stage one training from scratch. We found that the VPR performance is still maintained well on urban datasets, but reduces significantly on non-urban datasets, such as Nordland. This is likely due to the urban bias in common VPR pre-training datasets.

In the bottom four rows, we compared different configurations of \ourmethod{}. We observed that the performance of \ourmethod{} improves as we increase the encoder size. We further observed that increasing the number of top candidates passed to the pair classification component improves the recall, as it alleviates the performance limitation imposed by the global descriptor's effectiveness.

\newcount\columncount
\columncount = 10

\begin{table}[tb]
  \scriptsize
  \setlength\tabcolsep{0.1cm}
  \renewcommand{\arraystretch}{1.2}
  \centering
  \caption{Performance of the \ourmethod{} Global Descriptor.}%
  \resizebox{0.99\linewidth}{!}{
    \begin{tabular}{c|ccc||ccc||ccc}
\multirow{2}{*}{Method} & \multicolumn{3}{c||}{\textbf{MSLS-Val}}& \multicolumn{3}{c||}{\textbf{Pitts30k}}& \multicolumn{3}{c}{\textbf{Tokyo24/7}} \\
\cline{2-\columncount}
& R@1 & R@100 & R@500 & R@1 & R@100 & R@500 & R@1 & R@100 & R@500 \\
\cline{1-\columncount}
\noalign{\vskip\doublerulesep\vskip-\arrayrulewidth}
\cline{1-\columncount}
Pair-VPR-s (Global) & 86.4 & 97.8 & \underline{98.5} & 89.3 & 99.0 & \textbf{99.8} & 85.1 & \underline{99.1} & \underline{99.7} \\
Pair-VPR-s (Refine) & \underline{93.7} & 97.8 & 97.8 & \underline{94.7} & 99.0 & 99.0 & \underline{98.1} & \underline{99.1} & 99.1 \\
Pair-VPR-p (Global) & 86.6 & \underline{98.4} & \textbf{98.8} & 90.2 & \underline{99.1} & \underline{99.6} & 77.8 & \underline{99.1} & \textbf{100} \\
Pair-VPR-p (Refine) & \textbf{95.4} & \textbf{98.5} & \textbf{98.8} & \textbf{95.4} & \textbf{99.3} & \underline{99.6} & \textbf{100} & \textbf{100} & \textbf{100} \\
\end{tabular}%
    }
  \label{tab:global}%
  \vspace*{-0.3cm}
\end{table}%

\subsection{Performance of the Pair-VPR Global Descriptor}

In Table~\ref{tab:global}, we analyze the performance of the \ourmethod{} global descriptor. The performance of the second stage refinement is always limited by the recall of the global descriptor at the chosen number of top candidates, therefore we display Recall@N values for $N \in 1,100,500$. The global descriptor is always $512$ dimensions, allowing for computationally efficient database searching while also having a Recall@1 comparable to existing global descriptor VPR methods. We also note that our global descriptor is a lot smaller than the single-stage VPR methods in Table~\ref{tab:results}, \eg{} CosPlace, MixVPR and SALAD have $2048$, $4096$ and $8448$ dimensions respectively.

When we shift to using larger encoder sizes, we always maintain the same global descriptor by increasing the projection ratio from the class token. For a ViT-G encoder, the class token has $1530$ dimensions and we project down to $512$ dimensions using a linear layer. We do this in order to keep the computational cost of descriptor matching the same. We find that the ViT-G global descriptor performs similar to the ViT-B descriptor, except on Tokyo24/7. It is possible that the large projection ratio is not ideal for VPR, therefore future work should investigate experimenting with larger global descriptor sizes, albeit at an increased compute requirement.

\begin{figure}[tb]
    \centering
    \includegraphics[width=0.85\linewidth, trim=0cm 2.5cm 0cm 0.5cm, clip]{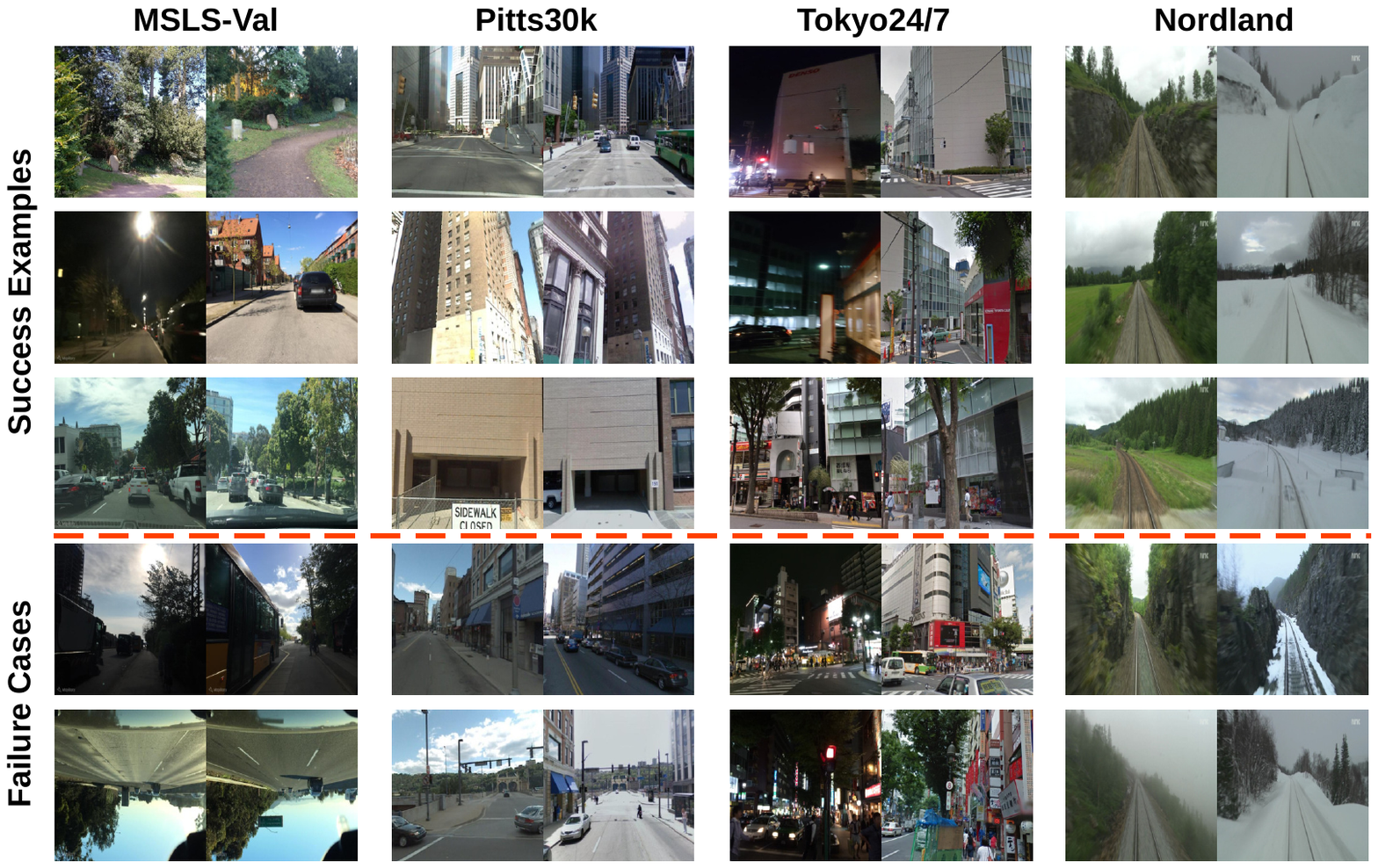}
    \caption{Qualitative results on the benchmark datasets, showing the performance of the computationally cheapest version of \ourmethod{} (Pair-VPR-s). We provide three success examples and two failure cases per dataset.}
    \label{fig:qual}
    \vspace{-0.4cm}
\end{figure}

\newcount\columncount
\columncount = 5

\begin{table}[b]
  \scriptsize
  \setlength\tabcolsep{0.1cm}
  \renewcommand{\arraystretch}{1.2}
  \centering
  \caption{Compute Analysis of two-stage methods.}%
  \resizebox{0.9\linewidth}{!}{
    \begin{tabular}{c|cccc}
\multirow{2}{*}{Method} & Encoding & Matching & Refinement & Storage \\
 & (ms/query) & (ms/query) & (ms/pair) & (mB/image) \\
\cline{1-\columncount}
\noalign{\vskip\doublerulesep\vskip-\arrayrulewidth}
\cline{1-\columncount}
Patch-NetVLAD-s & 12.51 & 0.17 & 0.94 & 1.83 \\

Patch-NetVLAD-p & 211.0 & 0.26 & 32.32 & 45.22 \\

R2Former & \textbf{7.11} & 0.33 & \textbf{0.74} & \textbf{0.25} \\

SelaVPR & 9.71 & \textbf{0.08} & \underline{0.91} & \underline{0.46} \\

\textbf{Pair-VPR-s (ViT-B)} & \underline{7.31} & \underline{0.13} & 7.18 & 1.55 \\

\textbf{Pair-VPR-p (ViT-G)} & 33.17 & \underline{0.13} & 7.31 & 3.10 \\

\end{tabular}%
    }
  \label{tab:compute}%
  \vspace*{-0.4cm}
\end{table}

\subsection{Compute Analysis}

In this subsection we discuss the computational requirements of \ourmethod{} and compare against other two-stage VPR methods on MSLS-Val, using the same compute platform (a single GPU), as shown in Table~\ref{tab:compute}. It can be seen that Pair-VPR-s is fast at encoding and matching, taking $7$ milliseconds per query image to encode and only $0.13$ ms/query to match against the entire database using $512$ dimensional descriptors. Our second stage refinement is slower than SelaVPR and R2Former but much faster than Patch-NetVLAD-p, and requires $0.7$ seconds per query with $100$ top candidates. Comparing the ViT-B to ViT-G encoder sizes of \ourmethod{}, only the encoding time and storage requirement increases, since we maintain the same global descriptor size and the same decoder network size. The storage requirements of \ourmethod{} are primarily from the dense features required for refinement and Pair-VPR-s features are comparable in size to Patch-NetVLAD-s and SelaVPR.

\subsection{Qualitative Analysis}

In Figure~\ref{fig:qual}, we provide qualitative results for Pair-VPR-s, including success and failure examples. The success examples show that Pair-VPR can successfully localize even in challenging situations such as at nighttime, under viewpoint shifts, or seasonal changes like summer to winter. The failure cases provide insights about challenging situations for VPR systems. For example, on the MSLS-Val dataset we show a failure due to heavy occlusion, and a second failure due to an upside down image in the dataset. In the case of the occluded image (due to a bus), these situations merit the addition of sequential VPR methods (\eg{}~\cite{garg2021seqnet}) to include adjacent image frames that may not be occluded. Concerning the failure examples on the Tokyo24/7 dataset, nighttime conditions are a common failure mode in VPR systems due to the lack of visible features at nighttime - although we note that these failure cases are solved by the more powerful Pair-VPR-p system, which has significantly more parameters in its ViT-G encoder.

\section{Conclusion}

In summary, we present \ourmethod{}, a novel two-stage VPR method that relies upon a mask image modeling pre-training strategy to maximize performance. \ourmethod{} combines a small $512$ dimensional global descriptor for rapid database searching along with a slower second stage that only searches a set of potential database matches. We observed that \ourmethod{} achieved the highest Recall@1 score on all datasets we tested on, comparing against recent state-of-the-art methods in VPR literature - while only requiring an encoder and decoder of $86$ million parameters each. As the encoder size is scaled up to ViT-L ($307$M) and ViT-G ($1.1$B), our training recipe allows for continuing performance improvements as the network size grows. Given the high recalls attained by \ourmethod{}, we believe it prudent to investigate converting this method from a VPR technique to a loop closure module in a full SLAM system. In the future, we aim to examine the effectiveness of \ourmethod{} at performing loop closures in SLAM. 

\FloatBarrier

\balance{}
\bibliographystyle{IEEEtran}
\bibliography{main}

\begin{thebibliography}{10}
\providecommand{\url}[1]{#1}
\csname url@samestyle\endcsname
\providecommand{\newblock}{\relax}
\providecommand{\bibinfo}[2]{#2}
\providecommand{\BIBentrySTDinterwordspacing}{\spaceskip=0pt\relax}
\providecommand{\BIBentryALTinterwordstretchfactor}{4}
\providecommand{\BIBentryALTinterwordspacing}{\spaceskip=\fontdimen2\font plus
\BIBentryALTinterwordstretchfactor\fontdimen3\font minus \fontdimen4\font\relax}
\providecommand{\BIBforeignlanguage}[2]{{%
\expandafter\ifx\csname l@#1\endcsname\relax
\typeout{** WARNING: IEEEtran.bst: No hyphenation pattern has been}%
\typeout{** loaded for the language `#1'. Using the pattern for}%
\typeout{** the default language instead.}%
\else
\language=\csname l@#1\endcsname
\fi
#2}}
\providecommand{\BIBdecl}{\relax}
\BIBdecl

\bibitem{vidanapathirana2022logg3d}
K.~Vidanapathirana, M.~Ramezani, P.~Moghadam \emph{et~al.}, ``{LoGG3D-Net: Locally Guided Global Descriptor Learning for 3D Place Recognition},'' in \emph{IEEE Int. Conf. Robot. Autom.}, 2022, pp. 2215--2221.

\bibitem{herraez2024spr}
D.~C. Herraez, L.~Chang, M.~Zeller \emph{et~al.}, ``{SPR: Single-Scan Radar Place Recognition},'' \emph{IEEE Robot. Autom. Lett.}, 2024.

\bibitem{AR2018}
R.~Arandjelović, P.~Gronat, A.~Torii \emph{et~al.}, ``{NetVLAD: CNN} architecture for weakly supervised place recognition,'' \emph{IEEE Trans. Pattern Anal. Mach. Intell.}, vol.~40, no.~6, pp. 1437--1451, 2018.

\bibitem{chen2017deep}
Z.~Chen, A.~Jacobson, N.~S{\"u}nderhauf \emph{et~al.}, ``Deep learning features at scale for visual place recognition,'' in \emph{IEEE Int. Conf. Robot. Autom.}\hskip 1em plus 0.5em minus 0.4em\relax IEEE, 2017, pp. 3223--3230.

\bibitem{jegou2010aggregating}
H.~J{\'e}gou, M.~Douze, C.~Schmid \emph{et~al.}, ``Aggregating local descriptors into a compact image representation,'' in \emph{IEEE Conf. Comput. Vis. Pattern Recog.}, 2010, pp. 3304--3311.

\bibitem{radenovic2018fine}
F.~Radenovi{\'c}, G.~Tolias, and O.~Chum, ``Fine-tuning cnn image retrieval with no human annotation,'' \emph{IEEE Trans. Pattern Anal. Mach. Intell.}, vol.~41, no.~7, pp. 1655--1668, 2018.

\bibitem{ali2022gsv}
A.~Ali-bey, B.~Chaib-draa, and P.~Gigu{\`e}re, ``Gsv-cities: Toward appropriate supervised visual place recognition,'' \emph{Neurocomputing}, vol. 513, pp. 194--203, 2022.

\bibitem{lu2024towards}
F.~Lu, L.~Zhang, X.~Lan \emph{et~al.}, ``Towards seamless adaptation of pre-trained models for visual place recognition,'' in \emph{Int. Conf. Learn. Represent.}, 2024.

\bibitem{hausler2021patch}
S.~Hausler, S.~Garg, M.~Xu \emph{et~al.}, ``Patch-netvlad: Multi-scale fusion of locally-global descriptors for place recognition,'' in \emph{IEEE Conf. Comput. Vis. Pattern Recog.}, 2021, pp. 14\,141--14\,152.

\bibitem{wang2022transvpr}
R.~Wang, Y.~Shen, W.~Zuo \emph{et~al.}, ``{TransVPR: Transformer-based place recognition with multi-level attention aggregation},'' in \emph{IEEE Conf. Comput. Vis. Pattern Recog.}, 2022, pp. 13\,648--13\,657.

\bibitem{vidanapathirana2023spectral}
K.~Vidanapathirana, P.~Moghadam, S.~Sridharan \emph{et~al.}, ``{Spectral Geometric Verification: Re-ranking Point Cloud Retrieval for Metric Localization},'' \emph{IEEE Robot. Autom. Lett.}, vol.~8, no.~5, pp. 2494--2501, 2023.

\bibitem{gupta2023siamese}
A.~Gupta, J.~Wu, J.~Deng \emph{et~al.}, ``Siamese masked autoencoders,'' \emph{Adv. Neural Inform. Process. Syst.}, vol.~36, pp. 40\,676--40\,693, 2023.

\bibitem{weinzaepfel2022croco}
P.~Weinzaepfel, V.~Leroy, T.~Lucas \emph{et~al.}, ``{CroCo: Self-Supervised Pre-training for 3D Vision Tasks by Cross-View Completion},'' \emph{Adv. Neural Inform. Process. Syst.}, vol.~35, pp. 3502--3516, 2022.

\bibitem{berton2022rethinking}
G.~Berton, C.~Masone, and B.~Caputo, ``Rethinking visual geo-localization for large-scale applications,'' in \emph{IEEE Conf. Comput. Vis. Pattern Recog.}, 2022, pp. 4878--4888.

\bibitem{weyand2020google}
T.~Weyand, A.~Araujo, B.~Cao \emph{et~al.}, ``Google landmarks dataset v2-a large-scale benchmark for instance-level recognition and retrieval,'' in \emph{IEEE Conf. Comput. Vis. Pattern Recog.}, 2020, pp. 2575--2584.

\bibitem{thoma2020soft}
J.~Thoma, D.~P. Paudel, and L.~V. Gool, ``Soft contrastive learning for visual localization,'' \emph{Adv. Neural Inform. Process. Syst.}, vol.~33, pp. 11\,119--11\,130, 2020.

\bibitem{uy2018pointnetvlad}
M.~A. Uy and G.~H. Lee, ``Pointnetvlad: Deep point cloud based retrieval for large-scale place recognition,'' in \emph{IEEE Conf. Comput. Vis. Pattern Recog.}, 2018, pp. 4470--4479.

\bibitem{leyva2023data}
M.~Leyva-Vallina, N.~Strisciuglio, and N.~Petkov, ``Data-efficient large scale place recognition with graded similarity supervision,'' in \emph{IEEE Conf. Comput. Vis. Pattern Recog.}, 2023, pp. 23\,487--23\,496.

\bibitem{yu2019spatial}
J.~Yu, C.~Zhu, J.~Zhang \emph{et~al.}, ``Spatial pyramid-enhanced netvlad with weighted triplet loss for place recognition,'' \emph{{IEEE} Trans. Neural Netw. Learn. Syst.}, vol.~31, no.~2, pp. 661--674, 2019.

\bibitem{ali2023mixvpr}
A.~Ali-Bey, B.~Chaib-Draa, and P.~Giguere, ``Mixvpr: Feature mixing for visual place recognition,'' in \emph{IEEE Winter Conf. Appl. Comput. Vis.}, 2023, pp. 2998--3007.

\bibitem{li2023hot}
Z.~Li, C.~D.~W. Lee, B.~X.~L. Tung \emph{et~al.}, ``{Hot-NetVLAD}: Learning discriminatory key points for visual place recognition,'' \emph{IEEE Robot. Autom. Lett.}, vol.~8, no.~2, pp. 974--980, 2023.

\bibitem{zhu2023r2former}
S.~Zhu, L.~Yang, C.~Chen \emph{et~al.}, ``{R2Former: Unified retrieval and reranking transformer for place recognition},'' in \emph{IEEE Conf. Comput. Vis. Pattern Recog.}, 2023, pp. 19\,370--19\,380.

\bibitem{geoadapt2024}
J.~Knights, S.~Hausler, S.~Sridharan \emph{et~al.}, ``{GeoAdapt: Self-Supervised Test-Time Adaptation in LiDAR Place Recognition Using Geometric Priors},'' \emph{IEEE Robot. Autom. Lett.}, vol.~9, no.~1, pp. 915--922, 2024.

\bibitem{dosovitskiy2020image}
A.~Dosovitskiy, L.~Beyer, A.~Kolesnikov \emph{et~al.}, ``An image is worth 16x16 words: Transformers for image recognition at scale,'' in \emph{Int. Conf. Learn. Represent.}, 2021.

\bibitem{chen2020simple}
T.~Chen, S.~Kornblith, M.~Norouzi \emph{et~al.}, ``A simple framework for contrastive learning of visual representations,'' in \emph{International conference on machine learning}.\hskip 1em plus 0.5em minus 0.4em\relax PMLR, 2020, pp. 1597--1607.

\bibitem{grill2020bootstrap}
J.-B. Grill, F.~Strub, F.~Altch{\'e} \emph{et~al.}, ``Bootstrap your own latent-a new approach to self-supervised learning,'' \emph{Adv. Neural Inform. Process. Syst.}, vol.~33, pp. 21\,271--21\,284, 2020.

\bibitem{caron2021emerging}
M.~Caron, H.~Touvron, I.~Misra \emph{et~al.}, ``Emerging properties in self-supervised vision transformers,'' in \emph{IEEE Conf. Comput. Vis. Pattern Recog.}, 2021, pp. 9650--9660.

\bibitem{oquab2023dinov2}
M.~Oquab, T.~Darcet, T.~Moutakanni \emph{et~al.}, ``{DINO}v2: Learning robust visual features without supervision,'' \emph{Transactions on Machine Learning Research}, 2024.

\bibitem{bao2021beit}
H.~Bao, L.~Dong, S.~Piao \emph{et~al.}, ``{BE}it: {BERT} pre-training of image transformers,'' in \emph{Int. Conf. Learn. Represent.}, 2022.

\bibitem{he2022masked}
K.~He, X.~Chen, S.~Xie \emph{et~al.}, ``Masked autoencoders are scalable vision learners,'' in \emph{IEEE Conf. Comput. Vis. Pattern Recog.}, 2022, pp. 16\,000--16\,009.

\bibitem{zhou2021ibot}
J.~Zhou, C.~Wei, H.~Wang \emph{et~al.}, ``Image {BERT} pre-training with online tokenizer,'' in \emph{Int. Conf. Learn. Represent.}, 2022.

\bibitem{haghighat2024ropim}
M.~Haghighat, P.~Moghadam, S.~Mohamed \emph{et~al.}, ``{Pre-training with Random Orthogonal Projection Image Modeling},'' in \emph{Int. Conf. Learn. Represent.}, 2024.

\bibitem{dao2022flashattention}
T.~Dao, D.~Fu, S.~Ermon \emph{et~al.}, ``Flashattention: Fast and memory-efficient exact attention with io-awareness,'' \emph{Adv. Neural Inform. Process. Syst.}, vol.~35, pp. 16\,344--16\,359, 2022.

\bibitem{Musgrave2020PyTorchML}
K.~Musgrave, S.~J. Belongie, and S.-N. Lim, ``Pytorch metric learning,'' \emph{ArXiv}, vol. abs/2008.09164, 2020.

\bibitem{wang2019multi}
X.~Wang, X.~Han, W.~Huang \emph{et~al.}, ``Multi-similarity loss with general pair weighting for deep metric learning,'' in \emph{IEEE Conf. Comput. Vis. Pattern Recog.}, 2019, pp. 5022--5030.

\bibitem{berton2023eigenplaces}
G.~Berton, G.~Trivigno, B.~Caputo \emph{et~al.}, ``{Eigenplaces: Training viewpoint robust models for visual place recognition},'' in \emph{IEEE Conf. Comput. Vis. Pattern Recog.}, 2023, pp. 11\,080--11\,090.

\bibitem{izquierdo2023optimal}
S.~Izquierdo and J.~Civera, ``Optimal transport aggregation for visual place recognition,'' in \emph{IEEE Conf. Comput. Vis. Pattern Recog.}, 2024, pp. 17\,658--17\,668.

\bibitem{warburg2020mapillary}
F.~Warburg, S.~Hauberg, M.~Lopez-Antequera \emph{et~al.}, ``Mapillary street-level sequences: A dataset for lifelong place recognition,'' in \emph{IEEE Conf. Comput. Vis. Pattern Recog.}, 2020, pp. 2626--2635.

\bibitem{torii2013visual}
A.~Torii, J.~Sivic, T.~Pajdla \emph{et~al.}, ``Visual place recognition with repetitive structures,'' in \emph{IEEE Conf. Comput. Vis. Pattern Recog.}, 2013, pp. 883--890.

\bibitem{torii201524}
A.~Torii, R.~Arandjelovic, J.~Sivic \emph{et~al.}, ``24/7 place recognition by view synthesis,'' in \emph{IEEE Conf. Comput. Vis. Pattern Recog.}, 2015, pp. 1808--1817.

\bibitem{skrede2013nordlandsbanen}
S.~Skrede, ``Nordlandsbanen: minute by minute, season by season,'' 2013.

\bibitem{zaffar2021vpr}
M.~Zaffar, S.~Garg, M.~Milford \emph{et~al.}, ``Vpr-bench: An open-source visual place recognition evaluation framework with quantifiable viewpoint and appearance change,'' \emph{Int. J. Comput. Vis.}, vol. 129, no.~7, pp. 2136--2174, 2021.

\bibitem{ali2024boq}
A.~Ali-bey, B.~Chaib-draa, and P.~Gigu{\`e}re, ``Boq: A place is worth a bag of learnable queries,'' in \emph{IEEE Conf. Comput. Vis. Pattern Recog.}, 2024, pp. 17\,794--17\,803.

\bibitem{garg2021seqnet}
S.~Garg and M.~Milford, ``Seqnet: Learning descriptors for sequence-based hierarchical place recognition,'' \emph{IEEE Robot. Autom. Lett.}, vol.~6, no.~3, pp. 4305--4312, 2021.

\end{thebibliography}

\end{document}